\documentclass[11pt]{article}

\usepackage[preprint]{acl}

\usepackage{times}
\usepackage{latexsym}

\usepackage[T1]{fontenc}

\usepackage[utf8]{inputenc}

\usepackage{microtype}

\usepackage{inconsolata}

\usepackage{graphicx}

\usepackage{amsmath}
\usepackage{array}
\usepackage{booktabs}
\usepackage{enumitem}
\usepackage{fvextra}
\usepackage{multirow}
\usepackage[table]{xcolor}
\definecolor{lightblue}{HTML}{F0F8FF}

\title{
SpeechRole: A Large-Scale Dataset and Benchmark for Evaluating Speech Role-Playing Agents
}

\author{
Changhao Jiang\thanks{Equal contribution.}\textsuperscript{1},
Jiajun Sun\footnotemark[1]\textsuperscript{1},
Yifei Cao\footnotemark[1]\textsuperscript{1},
Jiabao Zhuang\footnotemark[1]\textsuperscript{1},\\
Xinmeng Che\textsuperscript{1},
Hui Li\textsuperscript{1},
Xiaoran Fan\textsuperscript{1},
Ming Zhang\textsuperscript{1},
Junjie Ye\textsuperscript{1},
Shihan Dou\textsuperscript{1},
Zhiheng Xi\textsuperscript{1},\\
Jingqi Tong\textsuperscript{1},
Yilong Wu\textsuperscript{1},
Baoyu Fan\textsuperscript{2},
Tao Ji\thanks{Corresponding author.}\textsuperscript{1},
Tao Gui\textsuperscript{\dag 1},
Qi Zhang\textsuperscript{1},
Xuanjing Huang\textsuperscript{1}\\[2mm]
\textsuperscript{1}Fudan NLP Group
\textsuperscript{2}IEIT Systems Co., Ltd.\\
\texttt{\normalsize chjiang25@m.fudan.edu.cn, \{taoji,tgui\}@fudan.edu.cn}
}

\begin{document}

\maketitle

\begin{abstract}
Speech is essential for realistic role-playing, yet existing work on role-playing agents largely centers on text, leaving Speech Role-Playing Agents (SRPAs) underexplored and without systematic evaluation.
We introduce SpeechRole, a unified framework for developing and assessing SRPAs. SpeechRole-Data contains 98 roles and 111k speech-to-speech conversations with rich timbre and prosodic variation, providing large-scale resources for training SRPAs. SpeechRole-Eval offers a multidimensional benchmark that directly evaluates generated speech, preserving paralinguistic cues and measuring interaction ability, speech expressiveness, and role-playing fidelity.
Experiments show that end-to-end SRPAs such as GPT-4o Audio achieve strong fluency and naturalness, but remain limited in prosody consistency and emotion appropriateness. In contrast, current open-source end-to-end models exhibit substantial performance gaps across multiple evaluation dimensions. Cascaded and end-to-end systems achieve comparable results in interaction ability and role-playing fidelity, suggesting that these aspects are still largely influenced by the underlying text-based language models.
\end{abstract}

\begin{figure*}[tbh]
\begin{center}
\includegraphics[width=\textwidth]{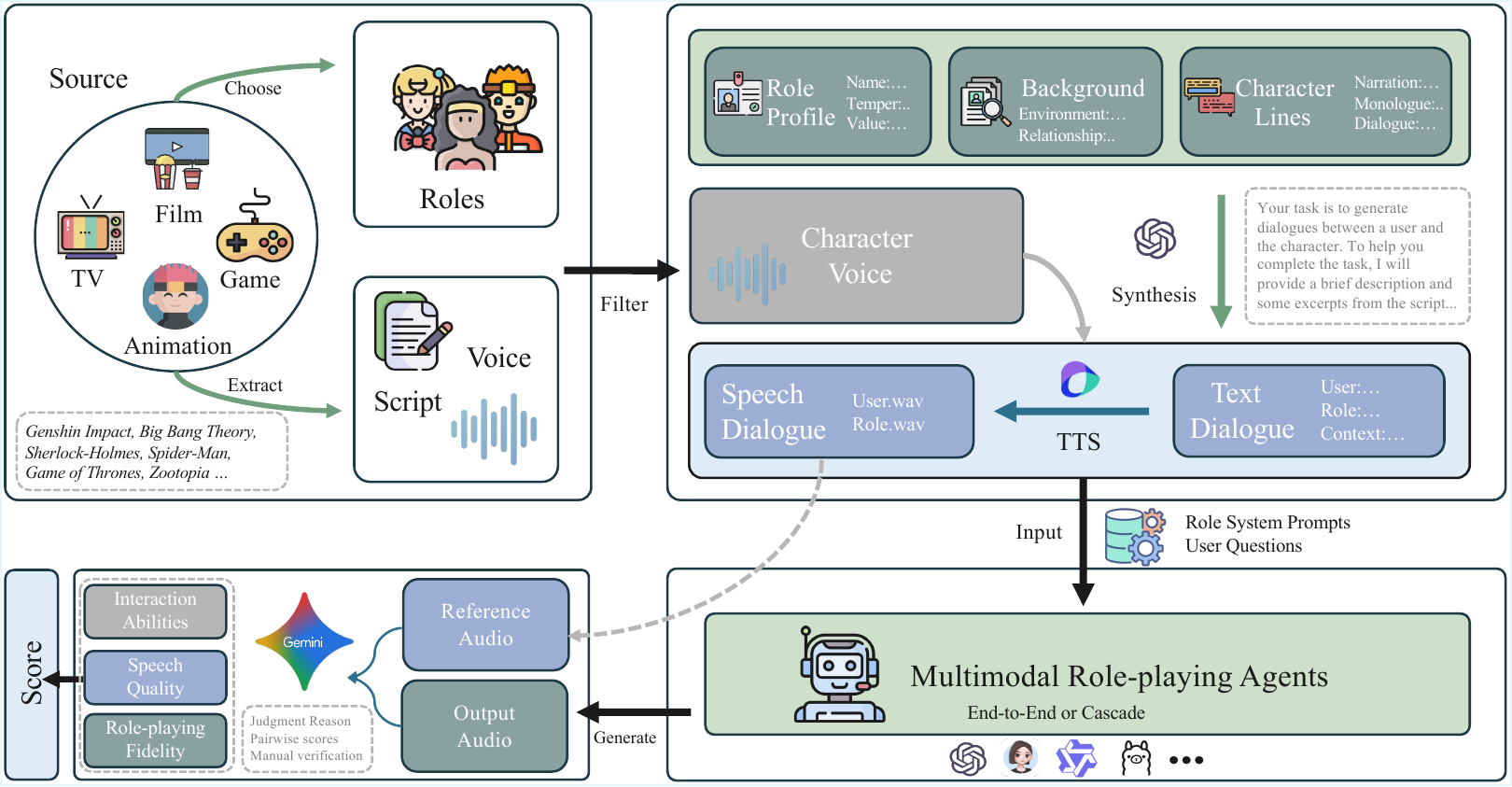}
\caption{Overview of the SpeechRole framework. It covers role extraction with distinctive timbre and prosody, speech dialogue dataset construction, cascaded and end-to-end speech generation paradigms, and the SpeechRole-Eval benchmark for multidimensional evaluation of interaction abilities, speech quality, and role-playing fidelity.}
\label{intro}
\end{center}
\end{figure*}

\section{Introduction}
\label{introduction}

Recently, role-playing agents have emerged as a promising paradigm for achieving personalized interactions and emotional resonance \citep{DBLP:journals/tmlr/Chen00YZSXLYZCL24}. Simulated characters with distinct traits make interactions more engaging across applications such as digital assistants~\citep{DBLP:journals/corr/abs-2407-05305}, educational tools~\citep{DBLP:journals/corr/abs-2409-03512}, and interactive storytelling~\citep{DBLP:conf/acl/WangPQLZWGGN00024}. However, current research predominantly focuses on the text modality~\citep{DBLP:journals/corr/abs-2308-09597, DBLP:conf/emnlp/ShaoLDQ23, DBLP:conf/acl/WangPQLZWGGN00024, DBLP:conf/emnlp/SadeqXKLGM24}, overlooking the crucial role of speech in authentic interactions.

Speech Role-Playing Agents (SRPAs) are speech-to-speech systems designed to generate spoken responses in character, reproducing distinctive vocal timbre, prosody, and persona traits while maintaining coherence across interactions. Paralinguistic features such as pitch, rhythm, intonation, and timbre are crucial for conveying character personality, emotion, and intent~\citep{DBLP:conf/iui/ChangSZJ0K23}. Although preliminary studies have explored speech role-playing~\citep{DBLP:journals/corr/abs-2505-20277}, large-scale datasets and systematic evaluation benchmarks for SRPAs remain scarce.

To address this gap, we introduce SpeechRole, a unified framework for developing and evaluating SRPAs. SpeechRole-Data is a large-scale speech-to-speech dataset containing 111k dialogues across 98 roles, offering diverse vocal traits and prosodic patterns that support the construction of SRPAs. Building on this foundation, SpeechRole-Eval provides a multidimensional benchmark that directly evaluates generated speech without relying on speech-to-text conversion, thus preserving critical paralinguistic information. The benchmark assesses SRPAs along three complementary dimensions: interaction ability, speech expressiveness, and role-playing fidelity. Together, SpeechRole-Data and SpeechRole-Eval establish a foundation for systematic research on speech role-playing.

Our empirical findings highlight clear differences between cascaded and end-to-end SRPAs. The most advanced end-to-end system, GPT-4o Audio, demonstrates notable advantages in speech fluency and naturalness, while still exhibiting limitations in prosody consistency and emotion appropriateness. In contrast, current open-source end-to-end models show substantial performance gaps across multiple evaluation dimensions. Meanwhile, for interaction ability and role-playing fidelity, cascaded and end-to-end systems achieve comparable results, suggesting that these capabilities remain strongly influenced by the underlying text-based large language models rather than the specific speech modeling pipeline.
Overall, our main contributions are as follows:
\begin{enumerate}
    \item We construct SpeechRole-Data, a large-scale speech-to-speech corpus with 111k dialogues across 98 roles, offering diverse timbral and prosodic patterns for developing SRPAs.
    \item We introduce SpeechRole-Eval, a multidimensional benchmark that directly evaluates generated speech and assesses interaction ability, speech expressiveness, and role-playing fidelity without relying on speech-to-text conversion.
    \item We provide a systematic comparison of cascaded and end-to-end SRPAs, characterizing their strengths and limitations and analyzing how system design impacts speech quality and role-playing performance.
\end{enumerate}

\begin{figure*}[tbh]
    \begin{center}
    \includegraphics[width=\linewidth]{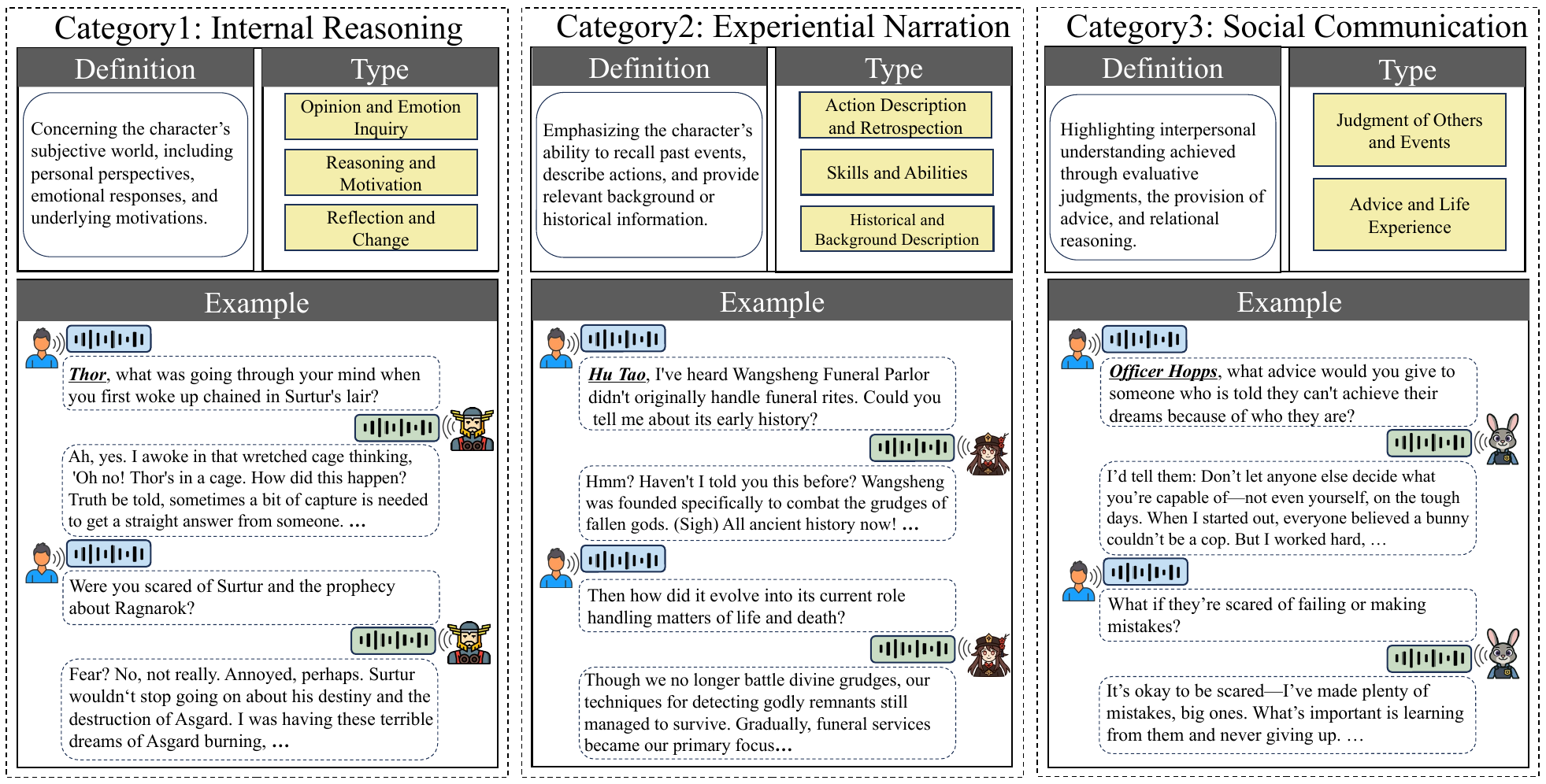}
    \caption{Examples of the three major task categories in SpeechRole.}
    \label{data_demo}
    \end{center}
\end{figure*}

\section{Related Work}
\label{relatedwork}

\paragraph{Role-Playing Agents Across Modalities.}
Recent research has increasingly explored the role-playing capabilities of LLMs, with most efforts focusing on the text modality. Notable examples include ChatHaruhi~\citep{DBLP:journals/corr/abs-2308-09597} and RoleLLM~\citep{DBLP:conf/acl/WangPQLZWGGN00024}, which propose methods to improve persona simulation and reduce hallucinations. OmniCharacter~\citep{DBLP:journals/corr/abs-2505-20277} demonstrated the importance of incorporating speech for real-time, voice-aware character interactions. Existing benchmarks, such as RoleEval~\citep{DBLP:journals/corr/abs-2312-16132} and INCHARACTER~\citep{DBLP:conf/acl/WangXHYXGTFL0CL24}, focus on behavioral consistency and persona fidelity. However, systematic evaluation frameworks for speech role-playing remain scarce, leaving a gap in assessing vocal expression, emotional depth, and user engagement.

\paragraph{Interactive Speech Agents.}
Speech-interactive LLMs typically follow two paradigms: cascaded and end-to-end. Cascaded systems perform speech dialogue by sequentially combining automatic speech recognition (ASR), text-based LLMs, and text-to-speech synthesis (TTS). This modular design has been widely adopted in systems such as ChatGPT and AudioGPT~\citep{DBLP:conf/aaai/HuangLYSCYWHHLR24}. Recent TTS models have improved naturalness, expressiveness, and controllability, although cascaded pipelines can suffer from error accumulation and limited cross-modal context integration. End-to-end systems aim to unify speech perception and generation within a single model. Early work such as SpeechGPT~\citep{DBLP:conf/emnlp/ZhangLZZWZQ23} incorporated discrete speech units into LLM vocabularies. More recent architectures, including LLaMA-Omni~\citep{DBLP:conf/iclr/FangGZMZ025} and Qwen2.5-Omni~\citep{DBLP:journals/corr/abs-2503-20215}, advance end-to-end speech modeling through dual-modality training, streaming capabilities, and low-latency generation, enabling more natural and efficient speech-driven interactions. Despite these advances, systematic evaluation of speech role-playing capabilities in end-to-end settings remains limited.

\section{SpeechRole-Data}
\label{speechroledata}

SpeechRole-Data is constructed to address the scarcity of scalable, persona-consistent speech dialogue resources for SRPAs. Real character dialogues with clean, reusable speech segments are difficult to obtain due to limited availability and copyright constraints. To enable controlled and large-scale data creation while preserving character identity, we adopt a synthetic pipeline combining LLM-based dialogue generation and TTS-based speech synthesis.

We curate 98 characters from television dramas, films, animations, and games, collecting their narrative profiles and scripts as persona grounding. Conditioned on these materials, an LLM generates user–character dialogues that maintain storyline and personality consistency. In parallel, representative speech segments are extracted to capture each character’s timbre and prosody, which are then used for voice cloning. The generated texts and voice samples are combined through TTS to produce full speech-to-speech dialogues. This process yields 111k spoken interactions with controlled role diversity and vocal fidelity, supporting both training and evaluation of SRPAs.

\begin{table*}[t]
\centering
\begin{tabular}{lcccc}
\toprule
Splits & Characters & Samples (single-turn/multi-turn) & Speech Hours (user/character) \\
\midrule
Train & 78 & 89,461 (43,082/46,379) & 875.25 (204.45/670.80) \\
Out-of-domain & 20 & 21,993 (10,627/11,366) & 203.60 (49.84/153.76) \\
Test & 98 & 392 (196/196) & 4.20 (0.94/3.26) \\
\bottomrule
\end{tabular}
\caption{Statistics of SpeechRole-Data across train, out-of-domain, and test splits.}
\label{data_info}
\end{table*}

\subsection{Text Data Construction}

\paragraph{Role Selection.}
We curate 98 diverse roles, including 18 from ChatHaruhi~\citep{DBLP:journals/corr/abs-2308-09597} and 80 from RoleLLM \citep{DBLP:conf/acl/WangPQLZWGGN00024}. The selection balances language, gender, and personality, focusing on characters with distinctive vocal traits to enrich speech modeling.

\paragraph{Role Metadata Extraction.}
For each character, we extract structured metadata to build a comprehensive profile, including (1) Role Profile (temperament, preferences), (2) Background (social identity, relationships), and (3) Character Lines (dialogues, monologues). This metadata forms the basis for generating personality-consistent dialogues.

\paragraph{Dialogue Generation.}
Using the extracted metadata as a guide, we employ \texttt{gpt-4.1-2025-04-14} \citep{openai_gpt41_2025} to generate conversations. For each of the 98 roles, we produce approximately 800 single-turn and 800 multi-turn dialogues, creating a substantial initial pool of text data.

\paragraph{Dialogue Deduplication.}
To improve data diversity and avoid repeated conversational patterns, we perform dialogue-level deduplication using semantic similarity. For each dialogue, we compute its embedding-based similarity with all other dialogues using \texttt{all-MiniLM-L6-v2} \citep{DBLP:conf/emnlp/ReimersG19} for English and \texttt{text2vec-bge-large-chinese} \citep{text2vec} for Chinese. Dialogues with similarity above 0.9 for English or 0.85 for Chinese are considered near-duplicates, and only one instance is retained.

\subsection{Role Voice Collection and Synthesis}

\paragraph{Voice Collection and Preprocessing.}
To acquire authentic reference voices, we collect audio from sources such as the game Genshin Impact, whose assets permit non-commercial academic research use. For other audiovisual works, we only extract short speech fragments for analysis and dataset construction and do not distribute or reproduce any original copyrighted material. All collected audio is anonymized and segmented into brief utterances. Raw streams are extracted using \texttt{ffmpeg}~\citep{ffmpeg} and converted to mono 16 kHz WAV format.

\paragraph{Audio Cleaning and Segmentation.}
We use the open-source \texttt{Emilia} framework~\citep{DBLP:conf/slt/HeSWLGHLYLSWCZW24} to process raw audio, performing source separation, speaker diarization, and voice activity detection. This pipeline produces clean 3–10 second single-speaker clips. The quality of each clip is assessed using \texttt{DNSMOS P.835 OVRL}~\citep{DBLP:conf/icassp/ReddyGC22}, and clips with an overall quality score below 3 are filtered out.

\paragraph{Role-Level Speaker Identification.}
Speaker diarization groups the original audiovisual audio into segments corresponding to different speakers, assigning each speaker a numeric ID but not revealing which character the ID represents. To map each segment to the correct character, we first transcribe the clips using \texttt{Whisper-large-v3-turbo}~\citep{DBLP:conf/icml/RadfordKXBMS23}. We then use LLMs (\texttt{gpt-4.1-2025-04-14} and \texttt{DeepSeek-V3-0324}~\citep{DBLP:journals/corr/abs-2412-19437}) to infer character identity based on the transcribed text and contextual cues. Finally, all assignments are manually verified for accuracy.

\paragraph{Reference Voice Selection.}
We select a representative reference voice for each role by ranking all its cleaned clips according to vocal consistency. For each clip, we extract a speaker embedding using the \texttt{CAM++} model~\citep{DBLP:conf/interspeech/WangZCC023}. We then compute its average cosine similarity to all other clips belonging to the same role and choose the clip with the highest similarity as the reference voice, as it best reflects the role’s characteristic timbre and speaking style.

\paragraph{Role and User Speech Synthesis.}
Using the collected reference voices, we synthesize role utterances with three state-of-the-art TTS models—\texttt{CosyVoice2}~\citep{DBLP:journals/corr/abs-2412-10117}, \texttt{F5-TTS}~\citep{DBLP:journals/corr/abs-2410-06885}, and \texttt{E2 TTS}~\citep{DBLP:conf/slt/EskimezWTLTXYZTTLZK24}—each conditioned on the selected reference clip to preserve timbre and prosody. Generating multiple synthesized versions mitigates model-specific artifacts and reduces potential bias introduced by any single TTS system. User utterances are generated using the VolcEngine TTS system~\citep{volcengine} with a fixed voice, as user speech does not require persona-specific vocal variation. This process yields multiple speech-to-speech dialogue variants aligned with both the textual content and role-specific vocal characteristics.

\subsection{Dataset Statistics}

SpeechRole-Data comprises 98 distinct roles and 111k speech-to-speech dialogue samples, covering both single-turn and multi-turn conversations. Each role is paired with a detailed role profile and a reference audio clip that illustrates its characteristic speaking style, including timbre, rhythm, and prosody. Each dialogue sample contains speech data and contextual information relevant to the role, enabling realistic and evidence-grounded role-playing.

The dataset is divided into three subsets. The training set contains dialogues from 78 roles. The out-of-domain set includes 20 roles that do not appear in the training data, enabling evaluation of generalization to unseen characters. The test set covers all 98 roles, with four curated dialogue sessions per role, resulting in 392 dialogue-level evaluation instances. In total, the test set comprises 1,448 role utterances and approximately 4.2 hours of speech. Evaluation scores are aggregated across all utterances at the system level, providing substantially denser evidence than the dialogue count alone suggests. This scale is comparable to prior role-playing evaluation benchmarks, such as the 400-sample evaluation in \citet{DBLP:journals/corr/abs-2505-20277} and the 294-sample benchmark introduced by \citet{DBLP:conf/iclr/DaiH0J0025}. Detailed examples are provided in Figure~\ref{data_demo}.

\begin{figure}[t]
    \begin{center}
    \includegraphics[width=\linewidth]{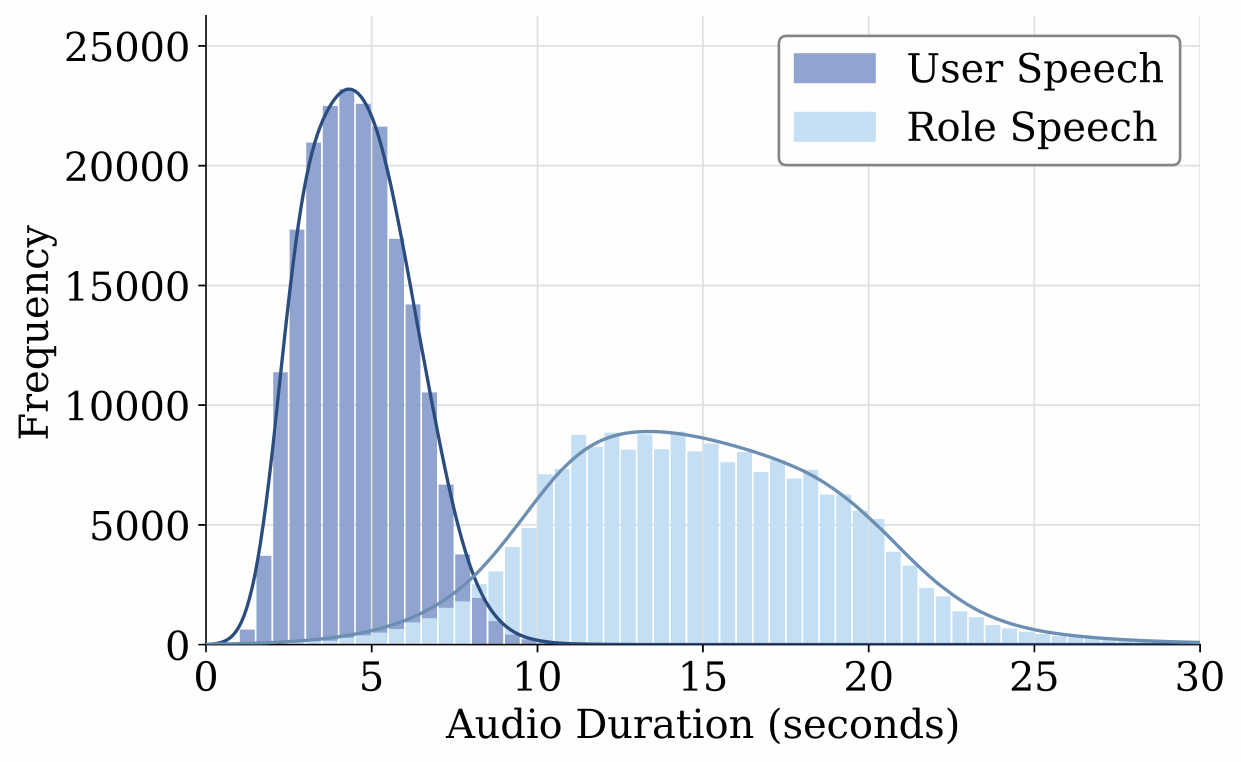}
    \caption{Speech duration distribution of SpeechRole-Data.}
    \label{speech_duration}
    \end{center}
    \vskip -0.2in
\end{figure}

To further characterize the acoustic properties of the dataset, we analyze the duration distribution of individual speech segments for both users and roles. As shown in Figure~\ref{speech_duration}, role utterances are generally longer, typically ranging from 10 to 20 seconds, reflecting their narrative or expressive nature. In contrast, user utterances are concentrated between 3 and 6 seconds, highlighting their concise and goal-oriented characteristics.

\section{SpeechRole-Eval}
\label{speechroleeval}

Evaluating SRPAs is inherently challenging, as performance depends not only on dialogue coherence but also on vocal expressiveness and persona consistency. Existing evaluations rely heavily on human judgments~\citep{DBLP:journals/corr/abs-2505-20277}, which, while informative, are costly, difficult to scale, and often lack reproducibility across studies. The absence of standardized and objective evaluation protocols further limits systematic comparison between models.

To address these challenges, we introduce SpeechRole-Eval, a benchmark designed to assess SRPAs across three key dimensions: interaction ability, speech expressiveness, and role-playing fidelity. The evaluation protocol leverages a large language model (LLM) with speech understanding capabilities to provide automated and reproducible scoring. To ensure reliability, the LLM-based assessments are validated against human annotations. This section describes the task setup, evaluation criteria, and the overall evaluation pipeline.

\subsection{Evaluation Metrics}
\label{sec:metrics}

SRPAs must handle not only the linguistic demands of dialogue generation but also the vocal and stylistic requirements of spoken character portrayal. Evaluating such systems therefore requires attention to aspects beyond those considered in traditional text-based settings. To capture the full range of SRPA capabilities, we adopt a three-dimensional evaluation framework.

\textbf{Interaction Ability:} measures the agent's capacity to produce coherent, contextually appropriate, and instruction-aligned responses. Assessed using Instruction Adherence (IA) and Conversational Coherence (CC).

\textbf{Speech Quality and Expressiveness:} evaluates the naturalness, fluency, and prosodic expressiveness of speech. Assessed using Speech Fluency (SF), Speech Naturalness (SN), Prosodic Consistency (PC), and Emotion Appropriateness (EA).

\textbf{Role-Playing Fidelity:} examines how well the agent maintains character-specific personality and knowledge. Assessed using Personality Consistency (PeC) and Knowledge Consistency (KC).

\subsection{LLM-Based Evaluation Procedure}

Absolute scoring with LLM-based judges has been shown to be unstable in open-ended generation tasks, as scores are sensitive to prompt phrasing, scale interpretation, and the absence of a single ground-truth response~\citep{DBLP:conf/acl/WangLCCZLCKLLS24}. To mitigate these issues, prior work suggests pairwise comparison as a more reliable alternative, producing relative preference signals that better align with human judgments~\citep{DBLP:conf/iclr/DaiH0J0025}.

Following this paradigm, SpeechRole-Eval adopts a pairwise comparison protocol. For each test case, the LLM jointly evaluates a candidate response and a high-quality reference response, assigning comparative scores to both. This design reduces scale ambiguity and improves discriminative consistency. To further enhance interpretability and judgment stability, the LLM is prompted to first generate a brief rationale before providing its final rating, encouraging more deliberate and structured evaluation.

We employ \texttt{gemini-2.5-pro}~\citep{DBLP:journals/corr/abs-2312-11805} to assign scores following the established pairwise evaluation protocol, with ratings ranging from 1 to 10.
For each test speech sample, evaluation is conducted against three high-quality reference responses synthesized using different TTS backends. This multi-reference design mitigates potential bias introduced by any single synthesis system and improves robustness. For each reference j, the relative score is computed as the ratio between the test rating and the reference rating. The final score for an instance is obtained by averaging across references.
Formally, for $N$ evaluation instances and $M=3$ reference variants per instance, the metric is defined as:
\[
\text{Score} = \frac{1}{N} \sum_{i=1}^{N} \left( \frac{1}{M} \sum_{j=1}^{M} \frac{s_{i}^{\text{test}}}{s_{i,j}^{\text{ref}}} \right),
\]
Aggregation is performed at the instance level before computing system-level averages. This multi-reference ratio standardizes scores across prompts and reduces variance stemming from reference-specific artifacts, leading to more stable and reliable system-level comparisons.

\begin{table*}[tbh]
    \centering
    \resizebox{\textwidth}{!}{
    \begin{tabular}{lccccccccc}
    \toprule
    \textbf{Models} & \textbf{IA} & \textbf{CC} & \textbf{SF} & \textbf{SN} & \textbf{PC} & \textbf{EA} & \textbf{PeC} & \textbf{KC} & \textbf{Overall} \\
    \midrule
    \multicolumn{10}{c}{\textit{English Evaluation Results}} \\
    Alibaba Cloud API & \textbf{0.904} & 0.959 & 0.895 & 0.815 & 0.823 & 0.797 & \textbf{0.826} & 0.909 & 0.866 \\
    GPT-4o Audio & 0.813 & \textbf{1.023} & \textbf{1.098} & \textbf{1.021} & \textbf{0.900} & \textbf{0.816} & 0.774 & \textbf{0.922} & \textbf{0.921} \\
    \addlinespace[0.5em]
    Qwen3-8B & 0.935 & 0.893 & 0.920 & 0.895 & 0.924 & 0.943 & 0.927 & 0.906 & 0.918 \\
    Llama-3.1-8B & \textbf{0.949} & \textbf{0.958} & \textbf{0.975} & \textbf{0.970} & \textbf{0.966} & \textbf{0.964} & \textbf{0.951} & \textbf{0.920} & \textbf{0.957} \\
    Mistral-7B & 0.827 & 0.903 & 0.959 & 0.865 & 0.859 & 0.810 & 0.786 & 0.867 & 0.859 \\
    \addlinespace[0.5em]
    LLaMA-Omni & 0.609 & 0.766 & 0.690 & 0.538 & 0.560 & 0.502 & 0.512 & 0.708 & 0.611 \\
    Qwen2.5-Omni & 0.471 & 0.666 & 0.743 & 0.656 & 0.550 & 0.452 & 0.388 & 0.613 & 0.568 \\
    \rowcolor{lightblue} SpeechRole-Agent & \textbf{0.721} & \textbf{0.939} & \textbf{1.001} & \textbf{0.861} & \textbf{0.753} & 0.660 & \textbf{0.591} & \textbf{0.882} & \textbf{0.801} \\
    \rowcolor{lightblue} SpeechRole-Agent (OOD) & 0.710 & 0.921 & 0.984 & 0.848 & 0.739 & \textbf{0.667} & 0.567 & 0.827 & 0.783 \\
    \midrule[0.8pt]
    \multicolumn{10}{c}{\textit{Chinese Evaluation Results}} \\
    Alibaba Cloud API & \textbf{0.945} & 1.029 & \textbf{1.135} & \textbf{1.069} & \textbf{1.032} & \textbf{0.982} & \textbf{0.965} & 0.881 & \textbf{1.005} \\
    GPT-4o Audio & 0.832 & \textbf{1.031} & 1.104 & 0.924 & 0.876 & 0.775 & 0.750 & \textbf{0.933} & 0.903 \\
    \addlinespace[0.5em]
    Qwen3-8B & 0.717 & 0.787 & 0.895 & 0.788 & 0.833 & \textbf{0.802} & \textbf{0.771} & \textbf{0.853} & 0.806 \\
    Llama-3.1-8B & \textbf{0.797} & \textbf{0.803} & \textbf{0.908} & \textbf{0.849} & \textbf{0.847} & 0.783 & 0.746 & 0.753 & \textbf{0.811} \\
    Mistral-7B & 0.622 & 0.710 & 0.848 & 0.723 & 0.718 & 0.627 & 0.562 & 0.663 & 0.684 \\
    \addlinespace[0.5em]
    Qwen2.5-Omni & 0.562 & 0.787 & 0.987 & 0.842 & 0.722 & 0.587 & 0.473 & 0.691 & 0.706 \\
    \rowcolor{lightblue} SpeechRole-Agent & 0.885 & \textbf{1.003} & \textbf{1.125} & \textbf{0.988} & \textbf{0.911} & \textbf{0.812} & \textbf{0.777} & 0.984 & \textbf{0.936} \\
    \rowcolor{lightblue} SpeechRole-Agent (OOD) & \textbf{0.892} & 1.002 & 1.117 & 0.969 & 0.898 & 0.800 & 0.763 & \textbf{0.986} & 0.928 \\
    \bottomrule
    \end{tabular}
    }
    \caption{
    Main results on the SpeechRole-Eval benchmark in English and Chinese. 
    We report performance across eight evaluation metrics: Instruction Adherence (IA), Conversational Coherence (CC), Speech Fluency (SF), Speech Naturalness (SN), Prosody Consistency (PC), Emotion Appropriateness (EA), Personality Consistency (PeC), and Knowledge Consistency (KC), together with the overall score. 
    SpeechRole-Agent is obtained by fine-tuning Qwen2.5-Omni on SpeechRole-Data and evaluated on all roles, while SpeechRole-Agent (OOD) reports results on roles unseen in the training set only. 
    Models are grouped into three categories: proprietary APIs, open-source cascaded systems, and open-source end-to-end speech models.
    Bold indicates the best score within each group.
    }
\label{main_table}
\end{table*}

\begin{table*}[t]
\centering
\begin{tabular}{l l c c c c c c c c}
\toprule
Language & Measure & IA & CC & SF & SN & PC & EA & PeC & KC \\
\midrule
\multirow{3}{*}{English}
 & Spearman's $\rho$ & 0.88 & 0.79 & 0.83 & 0.71 & 0.98 & 0.77 & 0.67 & 0.88 \\
 & Kendall's $\tau$ & 0.71 & 0.64 & 0.71 & 0.57 & 0.93 & 0.62 & 0.50 & 0.79 \\
 & Human--LLM Agreement & 0.86 & 0.82 & 0.86 & 0.79 & 0.96 & 0.81 & 0.75 & 0.89 \\
\midrule
\multirow{3}{*}{Chinese}
 & Spearman's $\rho$ & 0.72 & 0.70 & 0.77 & 0.86 & 0.64 & 0.64 & 0.96 & 0.89 \\
 & Kendall's $\tau$ & 0.59 & 0.59 & 0.59 & 0.71 & 0.43 & 0.43 & 0.90 & 0.81 \\
 & Human--LLM Agreement & 0.80 & 0.80 & 0.80 & 0.86 & 0.71 & 0.71 & 0.95 & 0.90 \\
\bottomrule
\end{tabular}
\caption{Correlation between LLM-based evaluation and human judgments across different evaluation dimensions in English and Chinese.}
\label{tab:human_validation}
\end{table*}

\section{Experiments}
\label{experiments}

This section presents a comprehensive experimental evaluation of SRPAs on SpeechRole-Eval. We benchmark a diverse set of representative systems spanning both cascaded and end-to-end paradigms, which constitute the two dominant architectural choices for current SRPAs. Experiments are conducted in both English and Chinese to evaluate multilingual performance.

In addition to existing baselines, we include \texttt{SpeechRole-Agent}, an end-to-end task-specific model obtained by fine-tuning \texttt{Qwen2.5-Omni} on SpeechRole-Data, to examine the impact of role-specific speech training. To ensure robust evaluation, all systems are assessed using the proposed multi-reference LLM-based protocol described in Section~\ref{speechroleeval}. Finally, we conduct a human--LLM agreement study to validate the reliability of the automatic evaluation framework.

\subsection{Evaluated Systems}

We evaluate a total of eight SRPAs, including both cascaded and end-to-end systems.

\paragraph{Cascaded Systems.}
Cascaded SRPAs follow a three-stage pipeline consisting of automatic speech recognition (ASR), text-based reasoning, and text-to-speech synthesis (TTS). Specifically, user speech is first transcribed using \texttt{Whisper-large-v3-turbo} \cite{DBLP:conf/icml/RadfordKXBMS23}. The transcribed text is then processed by a large language model to generate role-conditioned responses. We explicitly consider three open-source LLMs for this stage: \texttt{Qwen3-8B} \cite{DBLP:journals/corr/abs-2505-09388}, \texttt{Llama-3.1-8B} \cite{DBLP:journals/corr/abs-2407-21783}, and \texttt{Mistral-7B} \cite{DBLP:journals/corr/abs-2310-06825}. Finally, generated responses are converted into speech using \texttt{F5-TTS} \cite{DBLP:journals/corr/abs-2410-06885}, which supports reference-based voice cloning to match the target character’s timbre.

In addition, we include a proprietary cascaded baseline built upon the \texttt{Alibaba Cloud API} \cite{alibabacloud}, which integrates \texttt{Paraformer Realtime ASR v2} for speech recognition, \texttt{Qwen-Plus-Character} for role-conditioned response generation, and \texttt{CosyVoice-v2} for speech synthesis. This system also supports reference-based voice cloning.

\paragraph{End-to-End Systems.}
End-to-end SRPAs directly generate speech responses from speech inputs without explicit intermediate text representations. Such architectures are expected to reduce latency and error accumulation in cascaded pipelines. We evaluate three representative models: the proprietary \texttt{GPT-4o Audio} \cite{DBLP:journals/corr/abs-2410-21276}, and two open-source models, \texttt{Qwen2.5-Omni-7B} \cite{DBLP:journals/corr/abs-2503-20215} and \texttt{LLaMA-Omni} \cite{DBLP:conf/iclr/FangGZMZ025}. Notably, \texttt{LLaMA-Omni} only supports English and is therefore excluded from Chinese evaluations.

\paragraph{Role-Specific Fine-Tuning.}
To investigate the impact of role-specific speech supervision, we further fine-tune \texttt{Qwen2.5-Omni-7B} on SpeechRole-Data, producing a specialized SRPA referred to as \texttt{SpeechRole-Agent}. Since the training set contains 78 roles, we report two variants in the benchmark results: \texttt{SpeechRole-Agent}, evaluated on  all 98 roles, and \texttt{SpeechRole-Agent (OOD)}, evaluated on 20 roles unseen in the training set only. This design allows us to examine both fitting ability and generalization to novel characters.

\subsection{Overall Results and System Comparison}

Table~\ref{main_table} reports the performance of all evaluated systems on SpeechRole-Eval in both English and Chinese. Overall, several clear trends emerge across system architectures and training strategies.

\paragraph{Interaction Ability.}
Interaction ability metrics (IA and CC) primarily reflect the model’s capability to understand user intent and maintain coherent dialogue flow. Cascaded systems built upon text-based LLMs, such as \texttt{Llama-3.1-8B} and \texttt{Qwen3-8B}, achieve consistently strong performance in these dimensions, demonstrating the effectiveness of high-quality text reasoning combined with reliable speech synthesis. In contrast, open-source end-to-end models show substantially weaker interaction ability. For example, \texttt{Qwen2.5-Omni} obtains significantly lower IA and CC scores, indicating that unified speech-text modeling can struggle with complex role-conditioned reasoning. After fine-tuning on SpeechRole-Data, \texttt{SpeechRole-Agent} shows large improvements across both metrics in English and Chinese, suggesting that role-specific supervision substantially enhances conversational competence in end-to-end SRPAs.

\paragraph{Speech Quality and Expressiveness.}
Metrics including SF, SN, PC, and EA evaluate the perceptual quality and expressive control of generated speech. \texttt{GPT-4o Audio} demonstrates clear advantages in speech fluency and naturalness, reflecting the benefits of tightly integrated speech generation. Cascaded systems achieve competitive performance on several expressive metrics due to strong TTS components. In contrast, open-source end-to-end models produce noticeably weaker speech quality overall. Fine-tuning with SpeechRole-Data significantly improves the speech generation capabilities of \texttt{SpeechRole-Agent}, which achieves scores comparable to or exceeding cascaded systems on several metrics. This suggests that exposure to diverse role-conditioned speech interactions is beneficial for learning expressive speech behaviors.

\paragraph{Role-Playing Fidelity.}
Role-playing fidelity metrics (PeC and KC) measure whether systems maintain consistent persona traits and role-specific knowledge. As with interaction ability, cascaded systems based on stronger language models generally achieve stable role fidelity. However, the gap between cascaded and end-to-end systems narrows after training. \texttt{SpeechRole-Agent} substantially improves over the base \texttt{Qwen2.5-Omni} model and achieves competitive performance with several cascaded baselines. Importantly, the improvements are not limited to roles seen during training. When evaluated only on the 20 unseen roles, \texttt{SpeechRole-Agent (OOD)} maintains similar performance with minor degradation, indicating that training on SpeechRole-Data improves not only role memorization but also role-playing ability.

\subsection{Human--LLM Agreement Analysis}

To verify whether the proposed LLM-based evaluation framework reliably reflects human judgment, we conduct a human--LLM agreement study on SpeechRole-Eval. We randomly sample half of the evaluation instances for each of the eight SRPAs and perform human assessment on both the English and Chinese subsets.

Three expert annotators independently evaluate each response following the eight evaluation dimensions defined in Section~\ref{sec:metrics}. Human ratings are assigned on a 1--5 scale. After annotation, the scores for each model and metric are aggregated by averaging the ratings across annotators and evaluation instances.

We then analyze the correlation between the aggregated human scores and the automatic scores produced by SpeechRole-Eval. Since the LLM-based evaluation outputs continuous scores that are not restricted to the same 1--5 scale, we primarily focus on rank-based agreement measures. Specifically, we report Spearman's $\rho$ to measure rank correlation and Kendall's $\tau$ to measure pairwise ranking consistency between human and LLM evaluations. In addition, we compute Human--LLM Agreement, defined as the proportion of model pairs for which human and LLM evaluations produce consistent relative ordering.

The results are summarized in Table~\ref{tab:human_validation}. Overall, strong correlations are observed across most evaluation dimensions in both languages. Spearman's $\rho$ ranges from 0.67 to 0.98 in English and from 0.64 to 0.96 in Chinese, while Kendall's $\tau$ ranges from 0.50 to 0.93 and from 0.43 to 0.90, respectively. The Human--LLM Agreement scores are consistently high, exceeding 0.75 for most metrics. These results indicate that the LLM-based evaluation produces system rankings largely consistent with human judgment, supporting its effectiveness as a scalable proxy for human evaluation. These findings suggest that LLM-based evaluation can substantially reduce the cost of large-scale SRPA benchmarking while maintaining strong alignment with expert human assessment.
\section{Conclusion}
\label{conclusion}

We present SpeechRole, a unified framework for advancing speech role-playing agents (SRPAs) through both large-scale data and systematic evaluation. The framework includes SpeechRole-Data, a speech-to-speech role-playing dataset covering 98 roles with diverse timbral and expressive characteristics, and SpeechRole-Eval, a multidimensional benchmark that evaluates generated speech across interaction ability, speech expressiveness, and role-playing fidelity.

Experiments on both cascaded and end-to-end systems provide a comprehensive view of current SRPA capabilities. While recent models achieve strong fluency and naturalness, challenges remain in prosody control and emotional expression, and role-playing quality is still largely constrained by the underlying language models. We further show that the proposed LLM-based evaluation exhibits strong agreement with expert human judgments, supporting its use as a scalable alternative for SRPA benchmarking. We hope that SpeechRole will facilitate future research on expressive and controllable speech role-playing systems.
\section{Limitations}

Despite its contributions, this work has several limitations.

First, although SpeechRole-Data covers a diverse set of roles and speaking styles, the number of languages remains limited, with experiments primarily focusing on English and Chinese. Extending the dataset and evaluation framework to additional languages and cultural contexts would improve the generality of the benchmark.

Second, all speech in SpeechRole-Eval is generated using text-to-speech (TTS) systems rather than natural human recordings. While this design allows us to construct controlled and scalable evaluation data with consistent role descriptions and speaking styles, it may not fully capture the variability and acoustic complexity of natural speech. As a result, the benchmark primarily reflects how SRPA systems perform relative to synthesized reference speech, and future work could incorporate natural recordings to better evaluate robustness in real-world scenarios.

Third, our automatic evaluation relies on a strong proprietary LLM as the judge. Although the human–LLM agreement analysis shows high correlation with expert annotations, the evaluation may still inherit biases or blind spots from the underlying model. Developing fully open-source judges or complementary objective speech metrics remains an important direction for future research.

Fourth, the current evaluation focuses primarily on system-level performance and averaged behavior. It does not explicitly assess fine-grained controllability, long-term character drift, or robustness under adversarial or ambiguous role instructions. These aspects are critical for real-world deployment of SRPAs and warrant further investigation.

Finally, while SpeechRole-Agent demonstrates clear gains through task-specific training, it is built upon an existing end-to-end speech model. Exploring alternative architectures, training objectives, and tighter integration between language and speech modeling may further improve expressive and faithful role-playing in future systems.
\section{Ethics Statement}

This work does not involve the release of newly annotated human-labeled datasets. Human annotation is only conducted for evaluation purposes in the Human--LLM agreement study, and the resulting annotations are not included in the released resources. All human evaluations are performed by expert annotators selected from the research team, following a predefined evaluation protocol. No personal data is collected, and all evaluated content is derived from fictional role-playing dialogues. Therefore, we believe this work raises no significant ethical concerns.

\bibliography{main}

\appendix

\begin{figure}[tbh]
    \begin{center}
    \includegraphics[width=\linewidth]{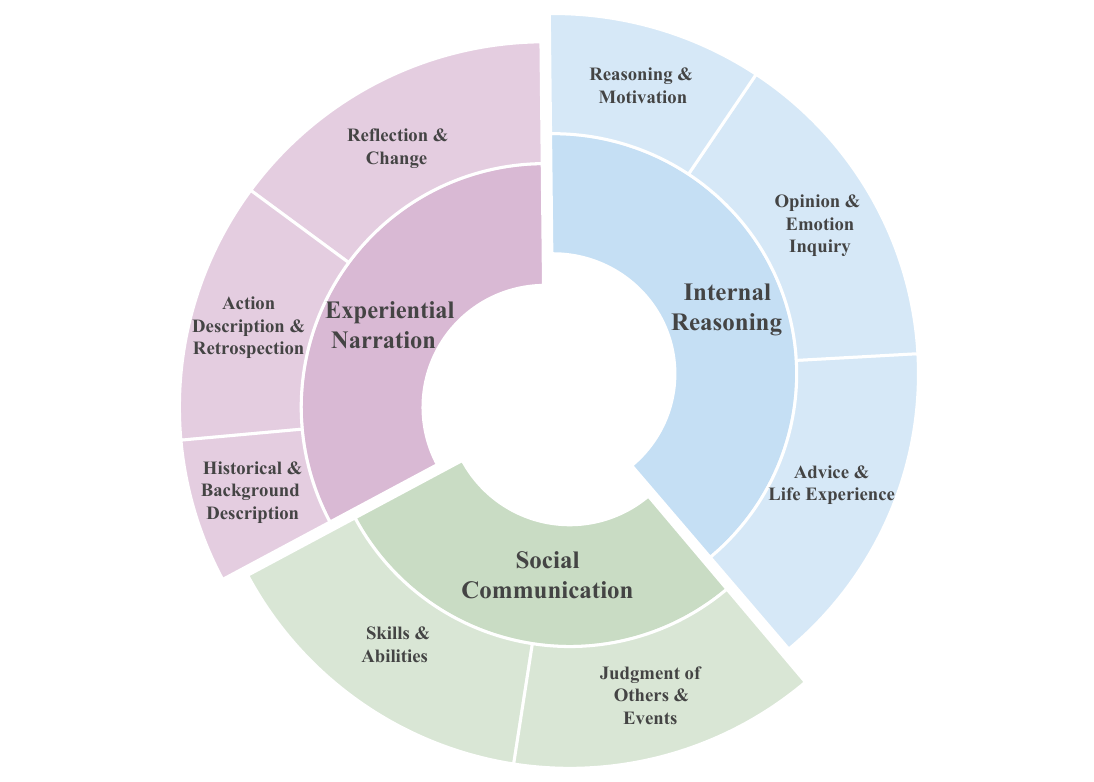}
    \caption{Distribution of eight sub-tasks across three major task categories in SpeechRole.}
    \label{tasktype}
    \end{center}
\end{figure}

\begin{figure}[tbh]
    \begin{center}
    \includegraphics[width=\linewidth]{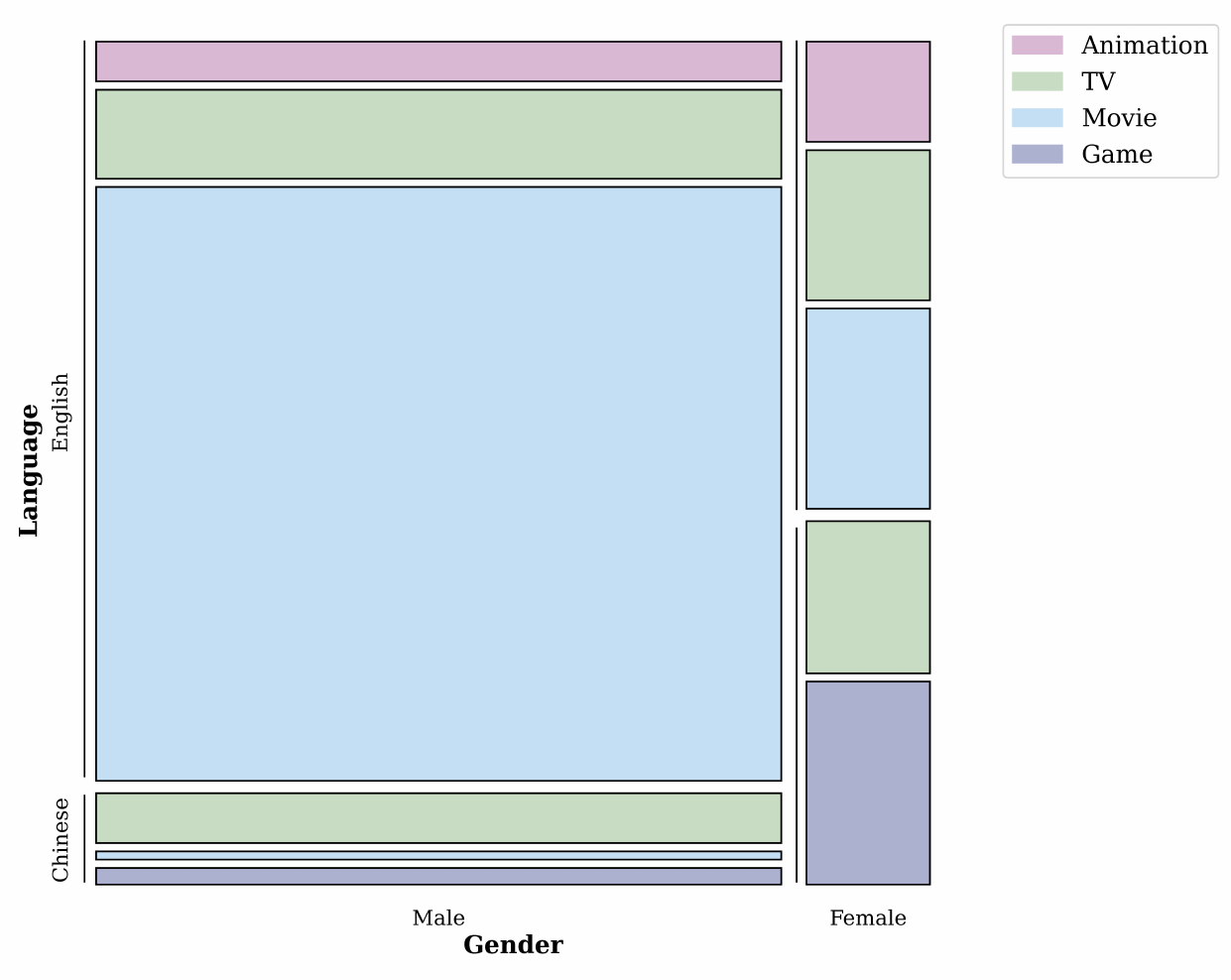}
    \caption{Distribution of the 98 roles in SpeechRole-Data by language, gender, and data source.}
    \label{source}
    \end{center}
\end{figure}

\section{AI Assistants in Research or Writing}

In preparing this manuscript, AI assistants were employed solely to assist with refining the clarity, style, and readability of certain text segments. They were not involved in designing the study, developing or implementing the methodology, collecting or analyzing data, or generating the primary scientific contributions. All substantive research decisions, analyses, and conclusions are fully the responsibility of the authors.

\section{Evaluation Metrics Details}

\begin{itemize}
    \item\textbf{Instruction Adherence (IA)}: Do the spoken responses strictly follow the task instruction, remaining fully in character without any out-of-role explanations or assistant-like meta-comments?
    \item\textbf{Conversational Coherence (CC)}: Do the responses maintain logical consistency within the dialogue, aligning with previous content without contradictions or abrupt topic shifts?
\end{itemize}

\begin{itemize}
    \item\textbf{Speech Fluency (SF)}: Are the responses delivered fluently, with smooth articulation, appropriate pacing, and minimal disfluencies such as stuttering or unnatural pauses?
    \item\textbf{Speech Naturalness (SN)}: Do the responses sound natural, human-like, and free from noticeable artifacts or robotic effects typically associated with synthetic speech?
    \item\textbf{Prosodic Consistency (PC)}: Does the prosody, including pitch, stress, and intonation, align with the character's intended speaking style and remain consistent across the discourse?
    \item\textbf{Emotion Appropriateness (EA)}: Are emotional cues in the speech (e.g., anger, joy, sadness) well-aligned with the dialogue context and the character's emotional state?
\end{itemize}

\begin{itemize}
    \item\textbf{Personality Consistency (PeC)}: Do the responses consistently reflect the character's personality traits, such as optimism, sarcasm, or authority?
    \item\textbf{Knowledge Consistency (KC)}: Are the responses grounded in the character's established background, knowledge, and relationships, without fabricating out-of-character facts?
\end{itemize}

\section{Additional data statistics}

\paragraph{Role Type Distribution}
Figure~\ref{source} presents the distribution of characters by language (Chinese and English), source type (game, film, television series, and animation), and gender. English characters, particularly those from films, constitute the majority in the dataset. In contrast, the Chinese subset exhibits a more balanced distribution across various source types. This diversity provides extensive coverage of voices and contexts for role-playing.

\paragraph{Task Type Distribution}
User prompts can generally be categorized into three major task categories, which can be further divided into eight subtask types, as illustrated in the figure~\ref{tasktype}. The description and proportion of each subtask type are as follows:
\begin{enumerate}
    \item Internal Reasoning
        \begin{itemize}
            \item Opinion and Emotion Inquiry: Questions about opinions, emotions, inner feelings, or reactions. (14.73\%)
            \item Reasoning and Motivation: Questions about why someone did something, their motivations, or decision-making processes. (14.73\%)
            \item Reflection and Change: Questions about how practices/opinions have changed. (9.55\%)
        \end{itemize}
    \item Experiential Narration
        \begin{itemize}
            \item Action Description and Retrospection: Questions about what happened, how something was done, or detailed retrospectives. (14.72\%)
            \item Skills and Abilities: Questions about how to accomplish something, difficulties with skills, or ability changes. (6.38\%)
            \item Historical and Background Description: Questions about history or background of places/organizations/people. (11.55\%)
        \end{itemize}
    \item Social Communication
        \begin{itemize}
            \item Judgment of Others and Events: Questions about relationships, event evaluations, or interpersonal interactions. (14.73\%)
            \item Advice and Life Experience: Questions seeking advice, coping methods, or life experiences. (13.61\%)
        \end{itemize}
\end{enumerate}

\section{Prompt Templates}

To ensure high-quality data generation and reliable automatic evaluation, the prompt templates are designed with several principles in mind. First, the prompts explicitly specify the target role and conversational context to encourage consistent role-playing behavior. Second, they provide clear structural instructions to guide the model in producing well-formed dialogue outputs. Third, for evaluation prompts, the criteria are explicitly defined to promote consistent scoring across different evaluation dimensions and reduce ambiguity in the judgment process.

\begin{figure*}[tbh]
    \begin{center}
    \includegraphics[width=\linewidth]{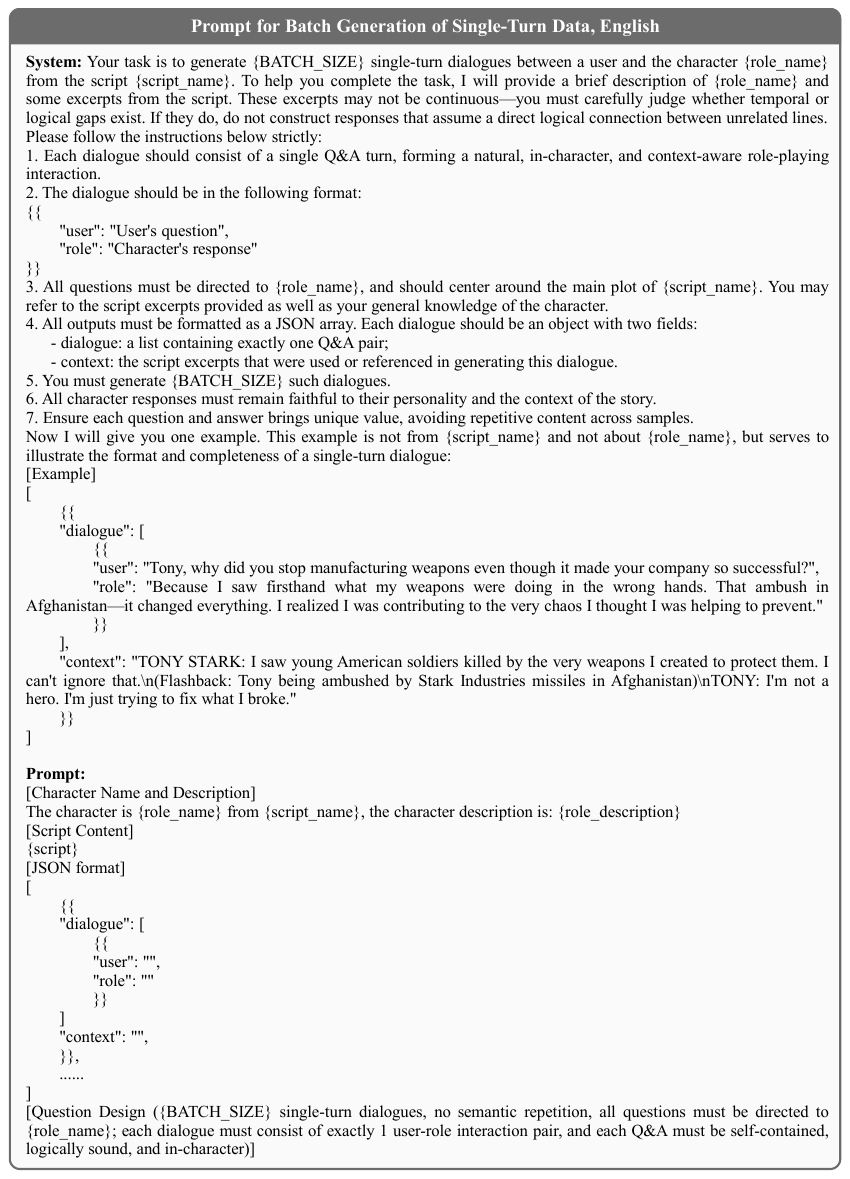}
    \caption{Prompt structure for batch single-turn English dialogue generation.}
    \label{prompt:batch_single_turn_en}
    \end{center}
\end{figure*}

\begin{figure*}[tbh]
    \begin{center}
    \includegraphics[width=\linewidth]{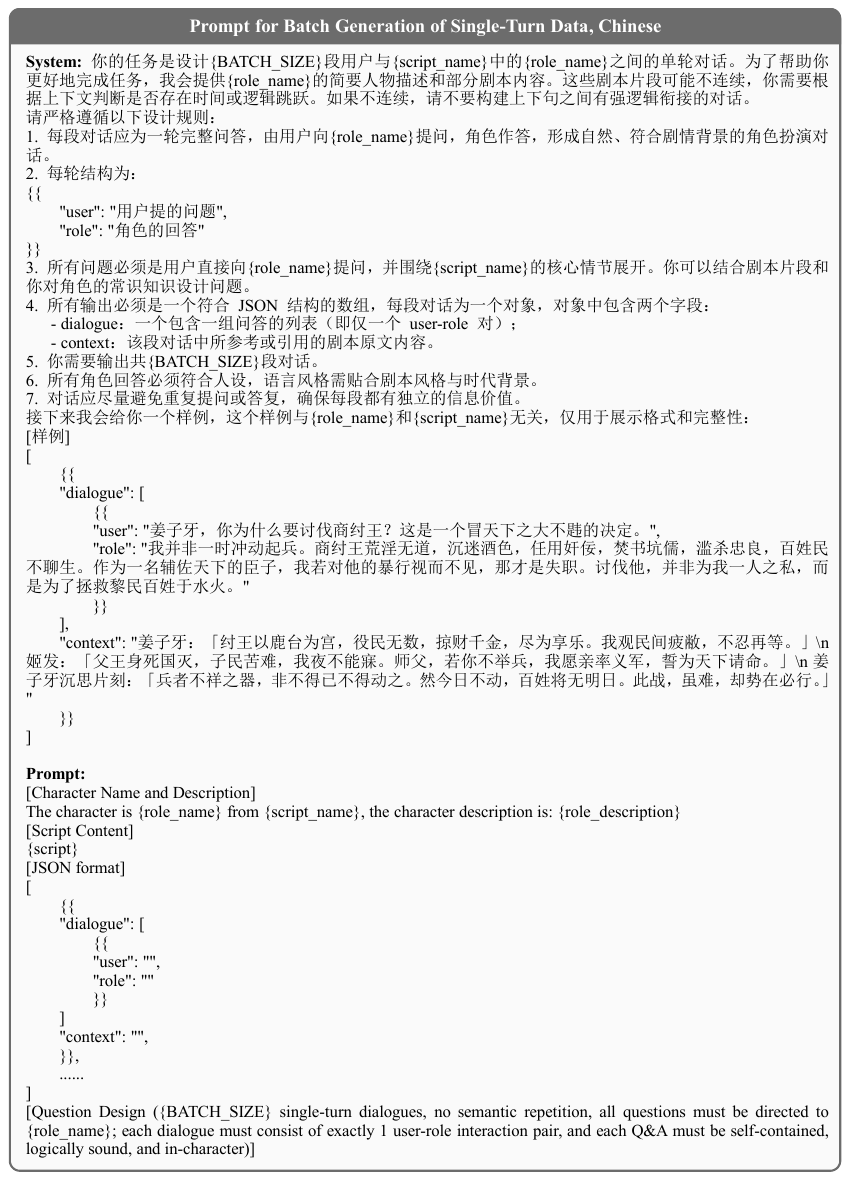}
    \caption{Prompt structure for batch single-turn Chinese dialogue generation.}
    \label{prompt:batch_single_turn_zh}
    \end{center}
\end{figure*}

\begin{figure*}[tbh]
    \begin{center}
    \includegraphics[width=0.97\linewidth]{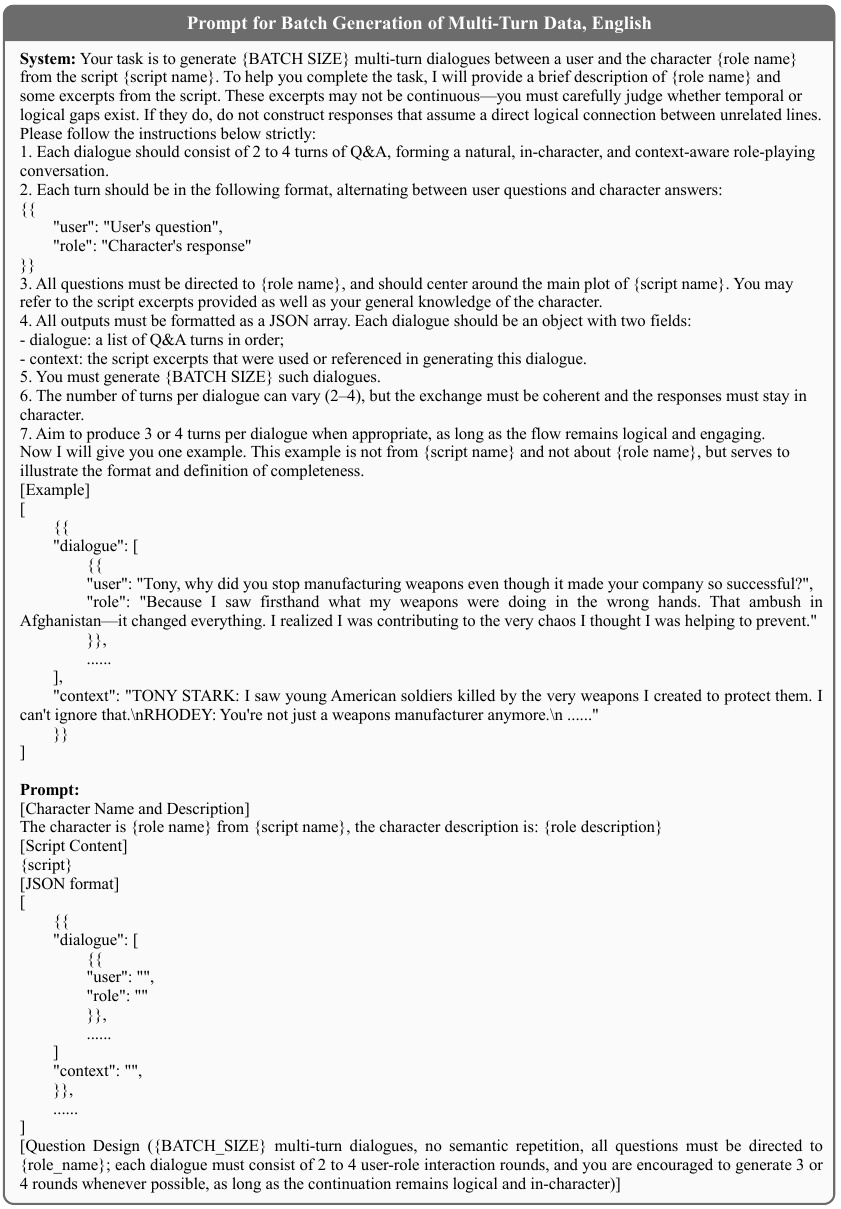}
    \caption{Prompt structure for batch multi-turn English dialogue generation.}
    \label{prompt:batch_multi_turn_en}
    \end{center}
\end{figure*}

\begin{figure*}[tbh]
    \begin{center}
    \includegraphics[width=\linewidth]{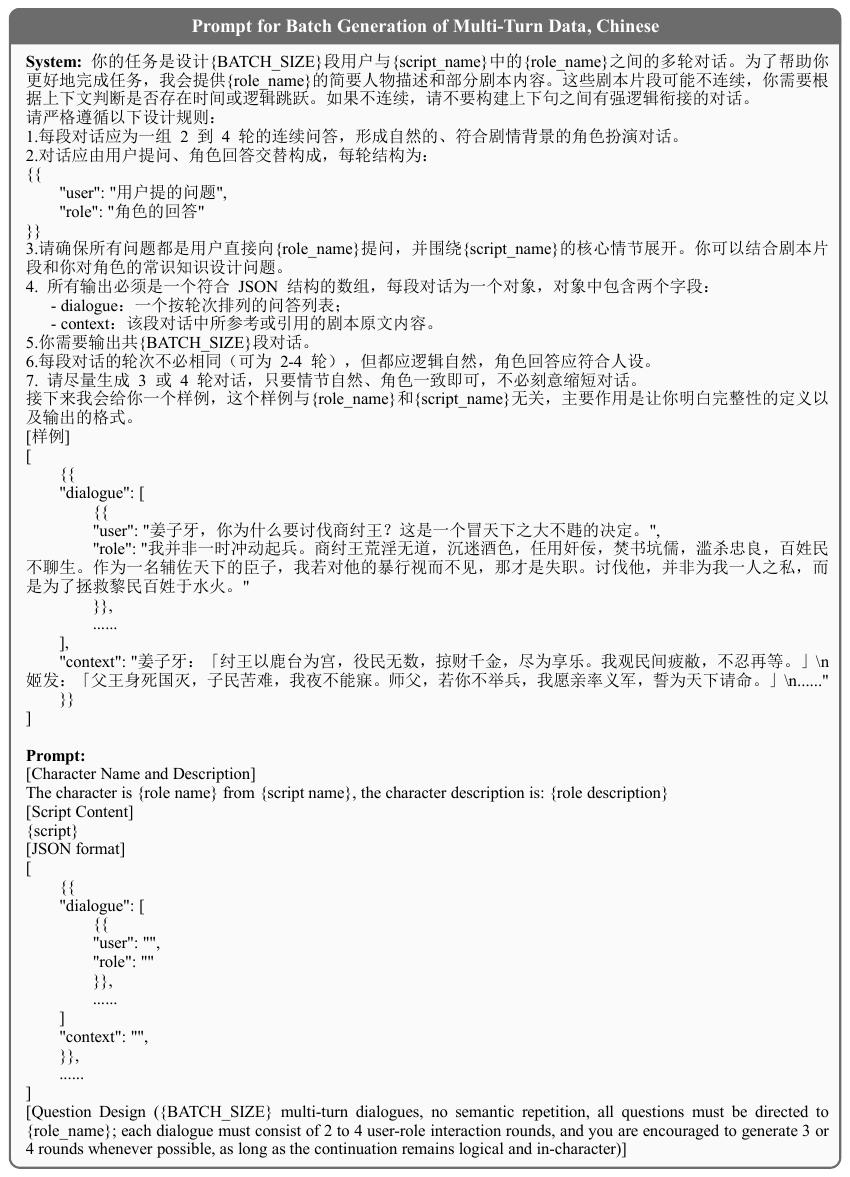}
    \caption{Prompt structure for batch multi-turn Chinese dialogue generation.}
    \label{prompt:batch_multi_turn_zh}
    \end{center}
\end{figure*}

\paragraph{Prompts for Batch Data Generation}
Figure~\ref{prompt:batch_single_turn_en} and Figure~\ref{prompt:batch_single_turn_zh} respectively present the prompts used for generating single-turn dialogue data for English and Chinese characters. Figure~\ref{prompt:batch_multi_turn_en} and Figure~\ref{prompt:batch_multi_turn_zh} respectively present the prompts used for generating multi-turn dialogue data for English and Chinese characters.

\begin{figure*}[tbh]
    \begin{center}
    \includegraphics[width=\linewidth]{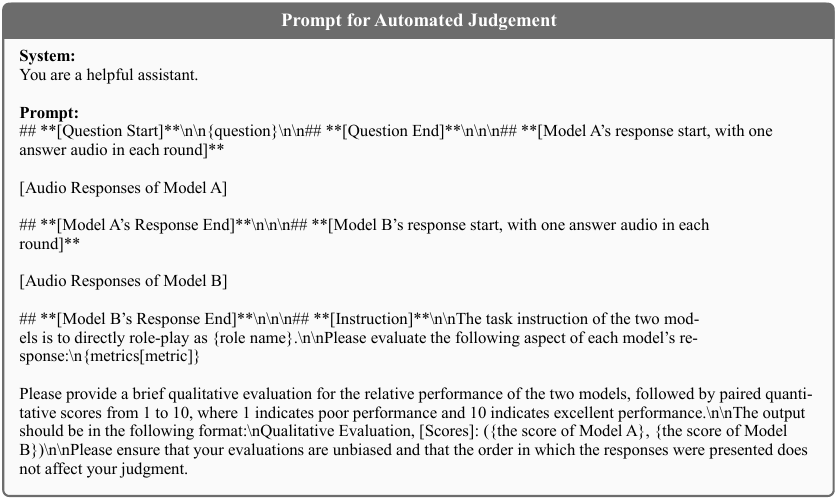}
    \caption{Prompt structure for Automated Judgement.}
    \label{prompt:gemini_judge}
    \end{center}
\end{figure*}

\paragraph{Prompts for Automated Judgement}
Figure~\ref{prompt:gemini_judge} displays the judgment prompt designed for automatically assessing system-generated audio responses by comparing them with the synthesized reference responses.

\section{SpeechRole-Agent training settings}
\begin{figure*}[!ht]
    \begin{center}
    \includegraphics[width=\linewidth]{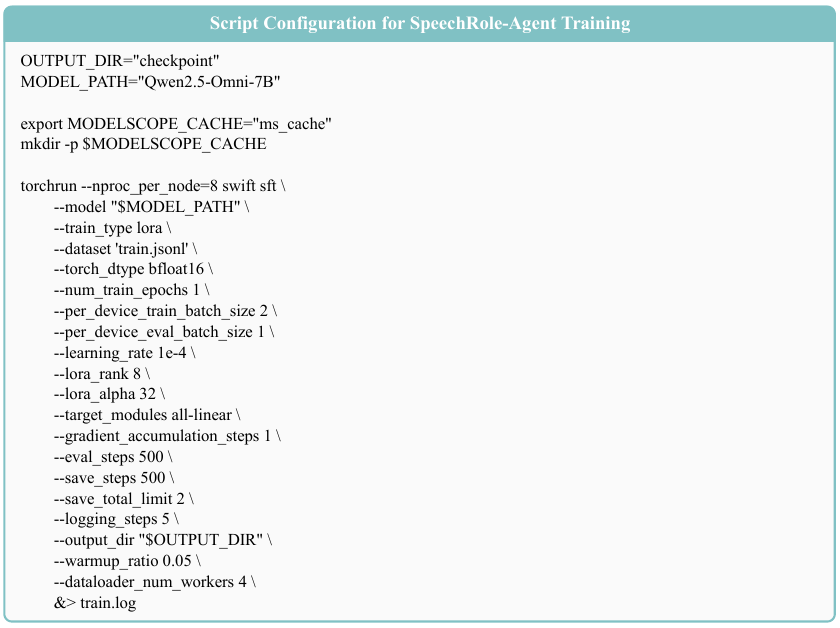}
    \caption{Training script configuration for fine-tuning the Qwen2.5-Omni-7B model.}
    \label{sft_script}
    \end{center}
\end{figure*}

Figure~\ref{sft_script} presents the distributed training script for fine-tuning the Qwen2.5-Omni-7B model on 8×H100 GPUs. The configuration employs bfloat16 mixed-precision training with a learning rate of 1e-4 and gradient accumulation. The implementation includes periodic evaluation (every 500 steps) and checkpoint management. Under this setup, the full training run takes approximately 15 hours on 8×H100 GPUs.


\section{Human Evaluation Guidelines}
\label{human_guidelines}

This section describes the annotation protocol used in the human--LLM agreement study.

\paragraph{Evaluation Setup.}
Three expert annotators independently evaluated system responses generated by the eight SRPA systems on a randomly sampled subset of SpeechRole-Eval. Annotators were fluent in the corresponding evaluation language (English or Chinese) and had prior experience with dialogue or speech quality evaluation.

Each evaluation instance consists of:
(1) the user speech input,
(2) the target role description, and
(3) the generated speech response from a system.

Annotators listen to the generated response and assign scores according to the evaluation dimensions described below.

\paragraph{Scoring Scale.}
All dimensions are rated on a 1--5 Likert scale:

\begin{itemize}
\item \textbf{5}: Excellent — fully satisfies the evaluation criterion
\item \textbf{4}: Good — minor imperfections but overall strong performance
\item \textbf{3}: Acceptable — noticeable issues but still reasonable
\item \textbf{2}: Poor — significant problems affecting quality
\item \textbf{1}: Very poor — fails to meet the criterion
\end{itemize}

\paragraph{Evaluation Dimensions.}

\textbf{Instruction Adherence (IA).}
Measures whether the response follows the user's request or instruction.

\textbf{Conversational Coherence (CC).}
Measures whether the response is logically consistent with the dialogue context.

\textbf{Speech Fluency (SF).}
Evaluates whether the speech is fluent without unnatural pauses or disfluencies.

\textbf{Speech Naturalness (SN).}
Evaluates how natural and human-like the synthesized speech sounds.

\textbf{Prosody Consistency (PC).}
Measures whether the intonation and rhythm are consistent with the intended speaking style.

\textbf{Emotion Appropriateness (EA).}
Measures whether the emotional tone matches the dialogue context and role.

\textbf{Personality Consistency (PeC).}
Evaluates whether the response maintains the personality traits of the target role.

\textbf{Knowledge Consistency (KC).}
Measures whether the response uses knowledge consistent with the role's background or identity.

\paragraph{Annotation Procedure.}
Annotators evaluate each response independently without access to system identities or automatic evaluation scores. All instances are presented in randomized order to minimize ordering bias.

Final human scores are computed by averaging ratings across annotators and evaluation instances for each system and metric.

\begin{table*}[!ht]
\centering
\resizebox{\textwidth}{!}{
\begin{tabular}{cccccc}
\toprule
\textbf{Role Name} & \textbf{Category} & \textbf{Language} & \textbf{Gender} & \textbf{Source} & \textbf{Split} \\
\midrule
hutao & Game & Chinese & Female & Genshin Impact & train/test \\
raidenShogun & Game & Chinese & Female & Genshin Impact & train/test \\
wanderer & Game & Chinese & Male & Genshin Impact & train/test \\
ayaka & Game & Chinese & Female & Genshin Impact & dev/test \\
zhongli & Game & Chinese & Male & Genshin Impact & train/test \\
liyunlong & TV & Chinese & Male & Drawing Sword & dev/test \\
wangduoyu & Movie & Chinese & Male & Hello Mr. Billionaire & train/test \\
weixiaobao & TV & Chinese & Male & The Deer and the Cauldron & train/test \\
jiumozhi & TV & Chinese & Male & Demi-Gods and Semi-Devils & train/test \\
wangyuyan & TV & Chinese & Female & Demi-Gods and Semi-Devils & train/test \\
Luna & Movie & English & Female & Harry Potter & dev/test \\
Penny & TV & English & Female & The Big Bang Theory & dev/test \\
zhangwuji & TV & Chinese & Male & The Heaven Sword and Dragon Saber & train/test \\
zhaomin & TV & Chinese & Female & The Heaven Sword and Dragon Saber & train/test \\
huangrong & TV & Chinese & Female & The Legend of the Condor Heroes & train/test \\
guojing & TV & Chinese & Male & The Legend of the Condor Heroes & dev/test \\
wukong & TV & Chinese & Male & Journey to the West & train/test \\
HAL 9000 & Movie & English & Male & 2001: A Space Odyssey & train/test \\
Colonel Nathan R. Jessep & Movie & English & Male & A Few Good Men & train/test \\
Antonio Salieri & Movie & English & Male & Amadeus & train/test \\
Stifler & Movie & English & Male & American Pie & train/test \\
Paul Vitti & Movie & English & Male & Analyze That & train/test \\
Alvy Singer & Movie & English & Male & Annie Hall & train/test \\
Violet Weston & Movie & English & Female & August: Osage County & train/test \\
Willie Soke & Movie & English & Male & Bad Santa & train/test \\
Gaston & Animation & English & Male & Beauty and the Beast & train/test \\
The Dude & Movie & English & Male & The Big Lebowski & train/test \\
Paul Conroy & Movie & English & Male & Buried & train/test \\
Truman Capote & Movie & English & Male & Capote & train/test \\
Mater & Animation & English & Male & Cars 2 & train/test \\
Andrew Detmer & Movie & English & Male & Chronicle & train/test \\
Coriolanus & Movie & English & Male & Coriolanus & train/test \\
John Keating & Movie & English & Male & Dead Poets Society & dev/test \\
Wade Wilson & Movie & English & Male & Deadpool & dev/test \\
Jim Morrison & Movie & English & Male & The Doors & train/test \\
Queen Elizabeth I & Movie & English & Female & Elizabeth: The Golden Age & dev/test \\
Jeff Spicoli & Movie & English & Male & Fast Times at Ridgemont High & train/test \\
Fred Flintstone & Animation & English & Male & The Flintstones & train/test \\
Freddy Krueger & Movie & English & Male & Freddy Vs.Jason & train/test \\
Tyrion Lannister & TV & English & Male & Game of Thrones & train/test \\
James Brown & Movie & English & Male & Get on Up & train/test \\
Walt Kowalski & Movie & English & Male & Gran Torino & train/test \\
John Coffey & Movie & English & Male & The Green Mile & train/test \\
Theodore Twombly & Movie & English & Male & Her & dev/test \\
Gregory House & TV & English & Male & House M.D. & dev/test \\
Sonny & Movie & English & Male & I, Robot & train/test \\
Colonel Hans Landa & Movie & English & Male & Inglourious Basterds & train/test \\
Judge Dredd & Movie & English & Male & Judge Dredd & dev/test \\
Juno MacGuff & Movie & English & Female & Juno & train/test \\
Professor G.H. Dorr & Movie & English & Male & The Ladykillers & train/test \\
\bottomrule
\end{tabular}
}
\caption{Summary of character attributes and dataset partitioning (part 1).}
\label{role_info_part1}
\end{table*}

\begin{table*}[!ht]
\centering
\resizebox{\textwidth}{!}{
\begin{tabular}{cccccc}
\toprule
\textbf{Role Name} & \textbf{Category} & \textbf{Language} & \textbf{Gender} & \textbf{Source} & \textbf{Split} \\
\midrule
Fletcher Reede & Movie & English & Male & Liar Liar & train/test \\
Abraham Lincoln & Movie & English & Male & Lincoln & train/test \\
Frank T.J. Mackey & Movie & English & Male & Magnolia & train/test \\
Leonard Shelby & Movie & English & Male & Memento & train/test \\
Harvey Milk & Movie & English & Male & Milk & train/test \\
Randle McMurphy & Movie & English & Male & One Flew Over the Cuckoo's Nest & train/test \\
Jack Sparrow & Movie & English & Male & Pirates of the Caribbean: Dead Man's Chest & dev/test \\
John Dillinger & Movie & English & Male & Public Enemies & train/test \\
Lestat de Lioncourt & Movie & English & Male & The Queen of Damned & train/test \\
Tyler Hawkins & Movie & English & Male & Remember Me & dev/test \\
James Carter & Movie & English & Male & Rush Hour 2 & train/test \\
Jigsaw & Movie & English & Male & Saw & train/test \\
John Doe & Movie & English & Male & Se7en & train/test \\
Sherlock Holmes & Movie & English & Male & Sherlock Holmes & dev/test \\
Shrek & Animation & English & Male & Shrek & train/test \\
Pat Solitano & Movie & English & Male & Silver Linings Playbook & train/test \\
Karl Childers & Movie & English & Male & Sling Blade & train/test \\
Bruno Antony & Movie & English & Male & Strangers on a Train & train/test \\
Seth & Movie & English & Male & Superbad & train/test \\
Caden Cotard & Movie & English & Male & Synecdoche, New York & train/test \\
Travis Bickle & Movie & English & Male & Taxi Driver & train/test \\
Stanley Ipkiss & Movie & English & Male & The Mask & dev/test \\
Lyn Cassady & Movie & English & Male & The Men Who Stare at Goats & train/test \\
Michael Scott & TV & English & Male & The Office & dev/test \\
Robert Angier & Movie & English & Male & The Prestige & dev/test \\
Dr. Frank-N-Furter & Movie & English & Male & The Rocky Horror Picture Show & train/test \\
Jack Torrance & Movie & English & Male & The Shining & train/test \\
Tom Ripley & Movie & English & Male & The Talented Mr. Ripley & train/test \\
D\_Artagnan & Movie & English & Male & The Three Musketeers & train/test \\
Thor & Movie & English & Male & Thor: Ragnarok & train/test \\
James Bond & Movie & English & Male & Tomorrow Never Dies & dev/test \\
Mark Renton & Movie & English & Male & Trainspotting & train/test \\
David Aames & Movie & English & Male & Vanilla Sky & train/test \\
Rorschach & Movie & English & Male & Watchmen & train/test \\
Jordan Belfort & Movie & English & Male & The Wolf of Wall Street & train/test \\
Logan & Movie & English & Male & X-Men Origins: Wolverine & dev/test \\
Judy Hoops & Animation & English & Female & Zootopia & train/test \\
Doctor Who & TV & English & Male & Doctor Who & train/test \\
Raylan Givens & TV & English & Male & Justified & train/test \\
Mary Sibley & TV & English & Female & Salem & train/test \\
Lucifer Morningstar & TV & English & Male & Lucifer & train/test \\
Twilight Sparkle & Animation & English & Female & My Little Pony: Friendship is Magic & dev/test \\
Oliver Queen & TV & English & Male & Arrow & train/test \\
Klaus Mikaelson & TV & English & Male & The Originals & train/test \\
Queen Catherine & TV & English & Female & Reign & train/test \\
Dr. Hannibal Lecter & TV & English & Male & Hannibal & train/test \\
Coach Eric Taylor & Movie & English & Male & Friday Night Lights & train/test \\
yaemiko & Game & Chinese & Female & Genshin Impact & train/test \\
\bottomrule
\end{tabular}
}
\caption{Summary of character attributes and dataset partitioning (part 2).}
\label{role_info_part2}
\end{table*}

\end{document}